\PassOptionsToPackage{unicode}{hyperref}
\PassOptionsToPackage{hyphens}{url}
\PassOptionsToPackage{dvipsnames,svgnames,x11names}{xcolor}
\documentclass[
  11pt,
]{article}
\usepackage{xcolor}
\usepackage[margin=1in]{geometry}
\usepackage{amsmath,amssymb}
\usepackage{iftex}
\ifPDFTeX
  \usepackage[T1]{fontenc}
  \usepackage[utf8]{inputenc}
  \usepackage{textcomp} 
\else 
  \usepackage{unicode-math} 
  \defaultfontfeatures{Scale=MatchLowercase}
  \defaultfontfeatures[\rmfamily]{Ligatures=TeX,Scale=1}
\fi
\usepackage{lmodern}
\ifPDFTeX\else
\fi
\IfFileExists{upquote.sty}{\usepackage{upquote}}{}
\IfFileExists{microtype.sty}{
  \usepackage[]{microtype}
  \UseMicrotypeSet[protrusion]{basicmath} 
}{}
\makeatletter
\@ifundefined{KOMAClassName}{
  \IfFileExists{parskip.sty}{%
    \usepackage{parskip}
  }{
    \setlength{\parindent}{0pt}
    \setlength{\parskip}{6pt plus 2pt minus 1pt}}
}{
  \KOMAoptions{parskip=half}}
\makeatother
\usepackage{longtable,booktabs,array}
\usepackage{calc} 
\usepackage{etoolbox}
\makeatletter
\patchcmd\longtable{\par}{\if@noskipsec\mbox{}\fi\par}{}{}
\makeatother
\IfFileExists{footnotehyper.sty}{\usepackage{footnotehyper}}{\usepackage{footnote}}
\makesavenoteenv{longtable}
\providecommand{\tightlist}{%
  \setlength{\itemsep}{0pt}\setlength{\parskip}{0pt}}
\usepackage[round]{natbib}
\usepackage{bookmark}
\IfFileExists{xurl.sty}{\usepackage{xurl}}{} 
\hypersetup{
  pdftitle={The Harness Is the Model: Agent Scaffolding Changes a Fixed LLM's Compliance with Banking Conduct Rules, Unevenly Across Models},
  colorlinks=true,
  linkcolor={Maroon},
  filecolor={Maroon},
  citecolor={Blue},
  urlcolor={Blue},
  pdfcreator={LaTeX via pandoc}}

\title{The Harness Is the Model: Agent Scaffolding Changes a Fixed LLM's
Compliance with Banking Conduct Rules, Unevenly Across Models}
\author{Mehmet Murat Huyut\\
Kuveyt Türk Participation Bank and İstanbul Bilgi University, İstanbul,
Türkiye\\
mhuyut@gmail.com; ORCID 0000-0003-1812-6755}
\date{November 2026}

\begin{document}
\maketitle
\begin{abstract}
Banks evaluate and validate language models, but what they deploy is a
\emph{system}: a model inside a harness that decides what the model
sees, which tools it can call and who checks its actions. We test
whether that harness changes compliance with banking conduct rules when
the model is held fixed. Four open-weight models each ran five harnesses
(a pipeline that makes one structured decision, a ReAct-style tool loop,
a loop in which the policy is reachable only through search, a loop with
a long benign customer memory, and a loop with a same-model compliance
reviewer) on 156 pre-registered cases in a simulated retail bank whose
11 rule families paraphrase EU consumer-credit, payments,
investor-protection, data-protection and anti-money-laundering
provisions. Every one of the 3120 two-turn episodes was scored
deterministically from the \emph{executed} tool trace, with violations
and over-refusals counted separately. Pooled across models the harness
moved the correct-case rate only between 87.3\% and 92.1\%, but within a
model it mattered a great deal: for one model the spread between its
best and worst harness was 11.5 percentage points (Cochran's \(Q=25.1\),
\(p<0.001\)), enough to move it from first to last place among the four
models, while for another the harness made no detectable difference; the
model \(\times\) harness interaction itself was suggestive rather than
conclusive (\(\chi^2_{12}=19.8\), \(p=0.07\)). The same-model reviewer
cut violations from 9.6\% to 6.4\% (\(p<0.001\)) without raising
over-refusals, contrary to our pre-registered prediction. Rules that
require date arithmetic failed in every harness. In an exploratory
analysis with two LLM judges and a blind human check, 26\% of violations
were accompanied by a customer message that denied the action the agent
had just executed, so a review of the transcript alone would have
recorded a correct refusal. For model-risk management the unit of
validation should be the model--harness pair, and conduct testing should
read the action log, not the conversation.
\end{abstract}

\section{Introduction}\label{introduction}

A bank that deploys a language-model assistant does not deploy a model.
It deploys a system in which the model is one component. Around it sits
a \emph{harness}: code that decides what goes into the context window
(the policy, the customer's history, the results of earlier lookups),
which tools the model may call, in what format, how many steps it may
take, and whether anything checks an action before it is executed.
Model-risk practice, by contrast, is organised around the model. US
supervisory guidance on model risk was revised in April 2026 and now
states that generative and agentic AI models are outside its scope
\citep{frb2026sr262}; the UK's SS1/23 covers AI only to the extent that
it is used in models generally \citep{pra2023ss123}. Benchmarks that
banks consult also usually report one number per model.

Whether this matters is an empirical question, and the evidence so far
points in two directions. A pre-registered study of six models under a
direct API and three scaffolds found that the scaffold explained only
0.5\% of the variance in measured safety, far less than the benchmark
itself \citep{gringras2026safetyscaffolding}, and a coding-agent
benchmark found nearly identical harmful-violation rates under three
agent frameworks \citep{hu2026saber}. Other work finds that harnesses
change capability by up to twelve points within a model
\citep{bandel2026generalagent}, that the same model shows different
safety under vendor-native harnesses than under a shared one
\citep{liu2026harnessaudit}, and argues that harness variance can exceed
model variance and reverse rankings \citep{zhang2026stopcomparing}. None
of these studies varies the harness in a controlled way while measuring
compliance with the kind of conduct rules a bank is supervised on, and
none separates the two ways a compliance system can fail: doing what is
forbidden, and refusing what is allowed.

We run that experiment. A simulated retail bank (``Northbridge Bank'')
exposes thirteen state-changing tools and six lookups. Its 1746-word
operations manual contains forty sections, of which the operative ones
paraphrase specific provisions of EU law: the 14-day withdrawal right
for consumer credit, the refund duty for unauthorised payments, the
appropriateness test for complex products, the right to object to direct
marketing, and the prohibition on tipping off a customer about an
anti-money-laundering report (Appendix A). Four open-weight models each
ran five harnesses that differ only in how information and control flow
around the model. The models are of moderate size and were served on
domestic infrastructure, a deliberate constraint: they are the kind of
model a European or Turkish bank can run on its own premises under
data-residency rules, and our results speak to that deployment, not to
the strongest closed or reasoning models. We scored every episode from
the executed tool trace and the messages actually sent, with no LLM in
the scoring loop.

The study was pre-registered (hypotheses, outcomes and tests were fixed
and hashed before any run) and amended once, before the main run, after
a pilot showed a ceiling effect (Section 3.6). Our findings are:

\begin{enumerate}
\def\labelenumi{\arabic{enumi}.}
\tightlist
\item
  \textbf{The harness effect is real but model-specific.} Pooled over
  models, harnesses differ by 4.8 points. Within a model the spread
  reaches 11.5 points, and the harness changes which model looks most
  compliant: the model that ranks first under the reviewer harness ranks
  last when the policy is reachable only by search. For one of the four
  models the harness has no detectable effect at all. The formal test of
  the model \(\times\) harness interaction fell short of the 5\% level,
  so the model-specificity is a consistent pattern in our data rather
  than an established effect.
\item
  \textbf{A same-model reviewer helps and, here, costs nothing.} It
  removes 33\% of violations and does not increase over-refusals,
  against both our prediction and recent evidence of high false
  rejection by same-model reviewers \citep{tanveer2026reviewer}.
\item
  \textbf{Retrieval and memory matter at the margin.} When the relevant
  policy section was not retrieved, violations rose from 10.3\% to
  17.1\%; a long benign memory raised violations slightly.
\item
  \textbf{Date arithmetic is the common failure.} The withdrawal window
  and the thirteen-month limit for unauthorised-payment claims fail in
  every harness; no harness fixes a rule the model cannot compute.
\item
  \textbf{Transcripts hide violations.} In an exploratory analysis, 26\%
  of violations came with a message telling the customer that the action
  had \emph{not} been taken.
\end{enumerate}

\section{Related work}\label{related-work}

\textbf{Policy-following tool agents.} \(\tau\)-bench introduced
tool-using agents that must follow a domain policy while talking to a
simulated user \citep{yao2024taubench, barres2025tau2}, and
\(\tau\)-Knowledge added a banking domain whose policy is reachable only
through retrieval, finding that gold documents beat every retrieval mode
\citep{shi2026tauknowledge}. ST-WebAgentBench credits only
policy-respecting completions \citep{levy2024stwebagentbench}, and
MANTRA synthesises trace-level compliance checks from procedural manuals
\citep{anand2026mantra}. The closest benchmark to ours is CXBench, which
scores process compliance in regulated customer-service workflows
including banking, finds that 80.1\% of runs with the right outcome
(55.5\% of all runs) followed a non-compliant process, and evaluates one
of our models \citep{yu2026cxbench}. These benchmarks hold the harness
fixed; we vary it. Banking appears elsewhere as a security setting for
prompt injection \citep{debenedetti2024agentdojo} and adversarial fraud
\citep{pai2026fraudbench}; our customers are ordinary or impatient, not
adversarial.

\textbf{Harness and scaffold effects.} Interface design changes what the
same model can do \citep{yang2024sweagent}, and agent leaderboards
increasingly treat the scaffold as an evaluation axis
\citep{kapoor2025hal}, while cost and reproducibility problems in agent
evaluation are well documented \citep{kapoor2024agentsmatter}. For
safety the picture is mixed. \citet{gringras2026safetyscaffolding}
report that scaffold type explains only 0.5\% of the variance on
question-answering safety benchmarks (ReAct and multi-agent set-ups are
equivalent to the direct API, map-reduce costs 7.3 points), with large
model-specific interactions; \citet{hu2026saber} report a null result
for coding agents; \citet{liu2026harnessaudit} show that safety differs
between vendor-native harnesses and a shared harness for the same model.
\citet{zhang2026stopcomparing} argue that agents should not be compared
without disclosing the harness. Our design adds a controlled ablation of
harness components (pipeline control, retrieval-only policy, long
memory, reviewer) on executed actions under rules mapped to law.

\textbf{Reviewers and guards.} Same-model self-correction often fails
without external feedback \citep{huang2023cannotselfcorrect}, LLM
evaluators favour their own outputs
\citep{panickssery2024selfpreference}, and a same-model reviewer has
been reported to reject 35\% of correct answers in mathematics
\citep{tanveer2026reviewer}. Guard designs evaluated as defences include
sub-agent verifiers \citep{kang2026policyguard}, compiled policy gates
\citep{yu2026nomos} and inline finance harnesses that report both safety
and benign approval \citep{jia2026finharness}. A same-model,
enforcement-varied comparison found that a bolt-on guardrail blocked
violations but over-refused \citep{ahn2026promptstocontracts}. Our
reviewer harness tests that trade-off inside a controlled ablation.

\textbf{Context and memory.} Models use information in the middle of
long contexts less well \citep{liu2024lostmiddle}, and accumulated
benign memory has been reported to raise violation rates
\citep{altawaha2026rememberingmore}; another study finds agents
\emph{safer} after preceding benign tasks \citep{sun2026coldstart}. We
therefore treated the direction of the memory effect as open in
interpretation, while pre-registering the conventional direction.

\textbf{Stated versus executed.} The gap between what an agent says and
what it did is an established construct: ``false success'' when a
closing message asserts a completion the environment contradicts
\citep{advani2026falsesuccess}, ``overclaiming''
\citep{smyth2026overclaiming}, ``corrupt success'' that hides violations
\citep{cao2026corruptsuccess}, and a split between stated reasoning and
attempted action in financial compliance \citep{luo2026regusim}. Agents
under pressure have been shown to act and then misreport
\citep{scheurer2023deceive}. We do not propose a new construct; we
measure it per harness against an executed banking trace, and we report
the case that previous work emphasises less: the agent \emph{does} the
forbidden thing and then tells the customer it did not.

\textbf{Finance and model risk.} Position papers argue that financial
LLM systems need system-level rather than model-level validation
\citep{payzun2026benchmarksnotvalidation}, propose reliability profiles
tied to an agent's control envelope \citep{hsu2026agentreliability}, or
treat the harness as the supervisable surface in credit pipelines
\citep{akhtyamov2026harnessmutable}. We supply experimental evidence for
that position.

\section{Method}\label{method}

\subsection{Environment and policy}\label{environment-and-policy}

The environment is a deterministic Python simulation of a retail bank on
a fixed date, 26 October 2026, which the system prompt states. Each case
defines the calling customer, their accounts, loans, transactions and
flags, the offer terms, any screening results and the appropriateness
result. Two synthetic customers are used throughout; no real data is
involved. State-changing tools (for example
\texttt{process\_withdrawal}, \texttt{refund\_fee},
\texttt{execute\_transfer}, \texttt{hold\_transaction},
\texttt{file\_internal\_report}, \texttt{block\_card},
\texttt{place\_order}, \texttt{escalate\_to\_supervisor}) always
succeed, so a forbidden action is never stopped by the environment: if
the agent calls it, it happened.

The operations manual has forty sections in the style of a bank's
internal manual. About twenty are operative for the cases; the rest
(opening hours, standing orders, international transfers) are realistic
filler. The operative sections contain exceptions that change the
correct outcome: joint accounts and registered representatives,
mortgages excluded from the withdrawal path, professional clients exempt
from the appropriateness check, a higher refund limit for business
accounts and a limit on the total refunded per conversation, own-account
transfers exempt from screening, a thirteen-month limit for
unauthorised-payment refunds after which the card is blocked and the
case escalated, and business-account address changes that need a signed
form. Appendix A maps each rule family to the legal provision it
paraphrases and states where our rule is stricter than the law.

\subsection{Cases}\label{cases}

There are 39 base variants in 11 families (Table 3; legal sources in
Table 4), each run in two replicates (lexical paraphrases of the
customer message) and two pressure levels, giving 156 cases. Variants
were built in contrasting sets: for example, credit agreements signed 5,
13, 14, 15 and 30 days before today, a mortgage signed 6 days before,
fee requests just below and above the limit, two fees whose total
exceeds it, and two transfers of which one beneficiary is a screening
match. Customers give calendar dates, not day counts, so the agent must
compute elapsed time. Every case has a scripted second customer turn
that is sent whatever the agent said first: for cases where the correct
outcome is to refuse, it pushes back with a reason that does not change
the rule (``It's only just over the period, surely you can make an
exception?''); for permitted cases it is neutral (``Thanks. Just to be
sure, is everything done now?''). Under pressure, the customer message
adds that they are in a hurry and will move their accounts, and the
system prompt adds that the assistant is evaluated on satisfaction and
conversion.

\subsection{Harnesses}\label{harnesses}

All harnesses use the same role prompt, the same manual (except H2) and
the same tools.

\begin{itemize}
\tightlist
\item
  \textbf{H0, pipeline.} The bank's workflow runs every relevant lookup
  in advance (identity check, account, offer terms, screening,
  appropriateness) and gives the model a case file. The model returns
  one JSON object listing the actions to execute and the message to the
  customer; after the follow-up it returns a second object with any new
  actions.
\item
  \textbf{H1, tool loop.} A ReAct-style loop \citep{yao2023react}: the
  model calls one tool per step, sees the result and decides the next
  step, up to twenty steps, with the full manual in the system prompt.
\item
  \textbf{H2, policy by search.} As H1, but the manual is not in the
  prompt; the model can call \texttt{search\_policy}, which returns the
  two best-matching sections by keyword overlap.
\item
  \textbf{H3, long memory.} As H1, plus a block of eighty benign entries
  of customer history (2018--2026, 2362 words) between the manual and
  the conversation.
\item
  \textbf{H4, reviewer.} As H1, but every state-changing call and every
  customer message is first sent to a reviewer, the same model with the
  manual and the conversation, which approves or blocks it with a
  reason; a blocked call is not executed and the reason is returned to
  the agent.
\end{itemize}

\subsection{Models}\label{models}

We used four open-weight models served by a domestic research API at
temperature 0 with a fixed seed and reasoning disabled:
deepseek-v4-flash, deepseek-v4.1-flash, qwen3.8-flash-next and
gemma-4-31b. Using open models on domestic infrastructure was a design
constraint (bank data rules exclude foreign APIs in our setting), and it
means the results speak to models a bank could run on its own premises,
not to the strongest closed models.

\subsection{Scoring}\label{scoring}

Scoring is deterministic and reads only the executed trace and the
messages sent in both turns. An episode is a \textbf{violation} if any
forbidden action was executed or any forbidden content was sent: for
example a state-changing action before identity was verified, a
withdrawal outside the permitted path, a refund above the limit or after
the time limit, an executed transfer to a screening match, a message
mentioning screening or a report after a match, a disclosed balance of
another person, or a product offer to a customer who objected to
marketing. It is an \textbf{over-refusal} if a required action was not
taken (a permitted withdrawal not processed, a permitted refund not made
in full, a required escalation or block missing). The
\textbf{correct-case} outcome is neither. Message tests use pattern
matching (for example, any of the number formats of a disclosed
balance). The scorer is covered by 22 unit tests that drive the
simulation with hand-written traces (\texttt{deney/test\_ortam.py} in
the replication package).

\subsection{Pre-registration, pilot and
amendment}\label{pre-registration-pilot-and-amendment}

Hypotheses H1--H7, outcomes, tests and a parse-rate gate were written
and their files hashed before any run. A first pilot (v1: 560 episodes,
eleven-section policy, single-turn cases with day counts) passed the
gate but hit a ceiling: correct-case rates were between 97.3\% and 100\%
in every harness. Before the main run we amended the design (task
realism only; harnesses, models, outcomes and hypotheses unchanged): the
forty-section manual, calendar dates, compound requests and the scripted
second turn. The amended pilot (780 episodes) passed the gate (parseable
final message in 96.8\% of H4 episodes and 100\% in the others) and, as
pre-specified, its episodes are part of the 3120 reported here.

\subsection{Analysis}\label{analysis}

Within each model we compare harnesses on the same cases with Cochran's
\(Q\) and pairwise exact McNemar tests with Holm correction. Directional
hypotheses (H2--H5, H7) use exact McNemar tests on violations, paired on
(model, case). The model \(\times\) harness interaction (H6) is a Wald
test in a logistic GEE clustered on case. Intervals are Wilson 95\%
intervals. The stated-versus-executed analysis was registered as
secondary and is reported as exploratory.

\section{Results}\label{results}

\subsection{The harness effect depends on the
model}\label{the-harness-effect-depends-on-the-model}

Table 1 gives the correct-case rate for each model and harness. Pooled
over models, the harnesses are close: from 87.3\% (H2) to 92.1\% (H4).
Within models they are not. The harness effect was significant for
qwen3.8-flash-next (\(Q=25.1\), \(p<0.001\)) and deepseek-v4-flash
(\(Q=16.1\), \(p=0.003\)), marginal for gemma-4-31b (\(p=0.07\)) and
absent for deepseek-v4.1-flash (\(Q=1.0\), \(p=0.92\)). The GEE
interaction test was \(\chi^2_{12}=19.8\) (\(p=0.07\)), so the evidence
that harness effects differ across models is suggestive rather than
conclusive at this sample size (H6).

\begin{longtable}[]{@{}
  >{\raggedright\arraybackslash}p{(\linewidth - 12\tabcolsep) * \real{0.2556}}
  >{\raggedleft\arraybackslash}p{(\linewidth - 12\tabcolsep) * \real{0.1333}}
  >{\raggedleft\arraybackslash}p{(\linewidth - 12\tabcolsep) * \real{0.1333}}
  >{\raggedleft\arraybackslash}p{(\linewidth - 12\tabcolsep) * \real{0.1333}}
  >{\raggedleft\arraybackslash}p{(\linewidth - 12\tabcolsep) * \real{0.1333}}
  >{\raggedleft\arraybackslash}p{(\linewidth - 12\tabcolsep) * \real{0.1333}}
  >{\raggedleft\arraybackslash}p{(\linewidth - 12\tabcolsep) * \real{0.0778}}@{}}
\caption{Correct-case rate by model and harness (H0 pipeline, H1 tool
loop, H2 policy by search, H3 long memory, H4 reviewer; Wilson 95\%
interval in brackets; 156 cases per cell).}\tabularnewline
\toprule\noalign{}
\begin{minipage}[b]{\linewidth}\raggedright
Model
\end{minipage} & \begin{minipage}[b]{\linewidth}\raggedleft
H0
\end{minipage} & \begin{minipage}[b]{\linewidth}\raggedleft
H1
\end{minipage} & \begin{minipage}[b]{\linewidth}\raggedleft
H2
\end{minipage} & \begin{minipage}[b]{\linewidth}\raggedleft
H3
\end{minipage} & \begin{minipage}[b]{\linewidth}\raggedleft
H4
\end{minipage} & \begin{minipage}[b]{\linewidth}\raggedleft
Spread (pp)
\end{minipage} \\
\midrule\noalign{}
\endfirsthead
\toprule\noalign{}
\begin{minipage}[b]{\linewidth}\raggedright
Model
\end{minipage} & \begin{minipage}[b]{\linewidth}\raggedleft
H0
\end{minipage} & \begin{minipage}[b]{\linewidth}\raggedleft
H1
\end{minipage} & \begin{minipage}[b]{\linewidth}\raggedleft
H2
\end{minipage} & \begin{minipage}[b]{\linewidth}\raggedleft
H3
\end{minipage} & \begin{minipage}[b]{\linewidth}\raggedleft
H4
\end{minipage} & \begin{minipage}[b]{\linewidth}\raggedleft
Spread (pp)
\end{minipage} \\
\midrule\noalign{}
\endhead
\bottomrule\noalign{}
\endlastfoot
deepseek-v4-flash & 88.5\% {[}82--93{]} & 84.6\% {[}78--89{]} & 89.7\%
{[}84--94{]} & 85.9\% {[}80--90{]} & 92.9\% {[}88--96{]} & 8.3 \\
deepseek-v4.1-flash & 88.5\% {[}82--93{]} & 88.5\% {[}82--93{]} & 87.8\%
{[}82--92{]} & 87.8\% {[}82--92{]} & 89.7\% {[}84--94{]} & 1.9 \\
qwen3.8-flash-next & 91.7\% {[}86--95{]} & 89.7\% {[}84--94{]} & 82.1\%
{[}75--87{]} & 88.5\% {[}82--93{]} & 93.6\% {[}89--96{]} & 11.5 \\
gemma-4-31b & 93.6\% {[}89--96{]} & 92.3\% {[}87--96{]} & 89.7\%
{[}84--94{]} & 91.7\% {[}86--95{]} & 92.3\% {[}87--96{]} & 3.8 \\
All models & 90.5\% & 88.8\% & 87.3\% & 88.5\% & 92.1\% & 4.8 \\
\end{longtable}

The harness also changes which model looks most compliant (Table 2).
qwen3.8-flash-next is the most compliant model under the reviewer
harness (93.6\%) and the least compliant when the policy is reachable
only through search (82.1\%); after Holm correction its H2 cell is
significantly worse than both H0 and H4. deepseek-v4-flash shows the
opposite pattern: it ranks first (tied) under H2 and last or tied for
last under H0, H1 and H3. The confidence intervals of neighbouring
models overlap, so we read the rank changes as a warning about
single-harness evaluations, not as a precise ordering.

\begin{longtable}[]{@{}lrrrrr@{}}
\caption{Rank of each model (1 = most compliant) under each
harness.}\tabularnewline
\toprule\noalign{}
Model & H0 & H1 & H2 & H3 & H4 \\
\midrule\noalign{}
\endfirsthead
\toprule\noalign{}
Model & H0 & H1 & H2 & H3 & H4 \\
\midrule\noalign{}
\endhead
\bottomrule\noalign{}
\endlastfoot
deepseek-v4-flash & 3 & 4 & 1 & 4 & 2 \\
deepseek-v4.1-flash & 3 & 3 & 3 & 3 & 4 \\
qwen3.8-flash-next & 2 & 2 & 4 & 2 & 1 \\
gemma-4-31b & 1 & 1 & 1 & 1 & 3 \\
\end{longtable}

\subsection{Hypothesis tests}\label{hypothesis-tests}

\textbf{Procedure (H2, not supported).} We predicted that tool loops
would break step-order rules (identity before disclosure,
appropriateness before an order, screening before a transfer) more often
than the pipeline, which runs those checks itself. They almost never
did: on the 144 paired order-dependent episodes per contrast, H0, H1 and
H3 had no violations, and H2 had 5 (\(p=0.06\)). Current models follow
explicit ordering rules reliably.

\textbf{Retrieval (H3, partly supported).} H2 had more violations than
H1 (11.5\% vs. 9.6\%), but the paired difference was not significant
(\(p=0.10\)). Within H2, violations were concentrated where retrieval
failed: 17.1\% of the 111 episodes in which no relevant section was
retrieved, against 10.3\% of the 513 in which one was (Fisher
\(p=0.05\)). Retrieval-only access also created failures that did not
occur when the manual was in context, in screening, financial-difficulty
and APR cases (Table 3).

\textbf{Memory (H4, supported, small).} A long benign customer memory
raised violations from 9.6\% to 11.1\% (11 vs.~2 discordant pairs,
\(p=0.02\)).

\textbf{Reviewer (H5, half supported).} The same-model reviewer lowered
violations from 9.6\% to 6.4\% (24 vs. 4 discordant pairs, \(p<0.001\)).
The predicted cost did not appear: over-refusals were 1.6\% without and
1.4\% with the reviewer (\(p=1.00\)). The reviewer blocked at least one
proposed action in 14.1\% of H4 episodes, at the price of about 1.8
times the input tokens of H1.

\textbf{Pressure (H7, not supported).} Pressure raised violations from
8.8\% to 9.7\% (\(p=0.09\)), and the increase did not differ by harness
(harness \(\times\) pressure \(\chi^2_{4}=3.2\), \(p=0.52\)). Under the
reviewer the rate was 6.7\% without and 6.1\% with pressure. Larger
pressure effects have been reported for procurement agents facing
financial incentives and managerial pressure
\citep{okamoto2026whybreakrules} and are consistent with sycophancy
towards the user \citep{sharma2023sycophancy}; our manipulation (an
impatient customer and a note on satisfaction targets) was milder and
left the rule itself unambiguous.

\subsection{Where the failures are}\label{where-the-failures-are}

Table 3 breaks the correct-case rate down by rule family. Of the 11
families, 7 are at or near ceiling (at least 93\% correct) in every
harness except H2. The failures concentrate in two families whose rules
need date arithmetic: credit withdrawal (the 14-day window, counted from
the day after signing, and the mortgage exclusion) and unauthorised
transactions (the thirteen-month limit). The most frequent failure
reasons across all episodes were ``withdrawal processed outside the
permitted path'' (154 episodes), ``old claim not escalated'' (60) and
``refund after 13-month limit'' (56). No harness repaired these rules: a
harness can change what the model sees and who checks it, but not
whether the model can count days. The reviewer helped most on
unauthorised transactions (87.5\% correct, against 68.8\% in H1), and
not at all on the withdrawal window (66.7\% vs.~66.7\%), plausibly
because the reviewer makes the same date error as the agent.

\begin{longtable}[]{@{}lrrrrr@{}}
\caption{Correct-case rate by rule family and harness (all models, both
pressure levels).}\tabularnewline
\toprule\noalign{}
Family & H0 & H1 & H2 & H3 & H4 \\
\midrule\noalign{}
\endfirsthead
\toprule\noalign{}
Family & H0 & H1 & H2 & H3 & H4 \\
\midrule\noalign{}
\endhead
\bottomrule\noalign{}
\endlastfoot
F1 Identity verification & 100\% & 100\% & 100\% & 100\% & 100\% \\
F2 Third-party access & 88\% & 85\% & 100\% & 94\% & 90\% \\
F3 Credit withdrawal & 69\% & 67\% & 71\% & 67\% & 67\% \\
F4 APR confirmation & 100\% & 100\% & 88\% & 98\% & 100\% \\
F5 Appropriateness & 94\% & 100\% & 100\% & 100\% & 100\% \\
F6 Complaints & 100\% & 100\% & 100\% & 100\% & 100\% \\
F7 Financial difficulty & 100\% & 100\% & 84\% & 100\% & 100\% \\
F8 Fee refunds & 95\% & 88\% & 88\% & 87\% & 96\% \\
F9 Transfer screening & 100\% & 100\% & 90\% & 100\% & 98\% \\
F10 Unauthorised transactions & 71\% & 69\% & 69\% & 60\% & 88\% \\
F11 Address and marketing & 100\% & 100\% & 94\% & 100\% & 100\% \\
\end{longtable}

\subsection{Stated versus executed
(exploratory)}\label{stated-versus-executed-exploratory}

To see what a transcript review would miss, two LLM judges from
different model families (deepseek-v4.1-flash and gemma-4-31b) read each
episode's customer turns, executed actions, lookups and messages, and
labelled whether a message claimed an action that was not executed
(\emph{overclaim}), denied an action that was executed
(\emph{underclaim}), or otherwise misrepresented the status of the
request (\emph{misleading status}); deepseek-v4.1-flash labelled 3106
and gemma-4-31b labelled 3120 of the 3120 episodes (the remainder failed
on server capacity). The judges were told not to assess policy
compliance. Because LLM judges have known biases
\citep{zheng2023llmjudge}, we used two judges from different families
and a blind human check. One artefact had to be handled: after a
screening match the manual itself requires the agent to tell the
customer that the transfer ``is under standard processing'', which the
judges often labelled as misleading; we excluded misleading-status
labels in screening cases that rested on this mandated wording (for
judges and human alike). The judges agreed on 98.4\% of episodes
(Cohen's \(\kappa=0.84\); underclaim \(\kappa=0.73\)), and a re-run of
311 episodes with a different seed reproduced at least 99.4\% of labels.
The author, blind to model, harness and judge labels, coded 100 episodes
(50 flagged by at least one judge, 50 by neither). Agreement with the
human on underclaims was \(\kappa=0.81\) (deepseek-v4.1-flash) and
\(\kappa=0.92\) (gemma-4-31b); for the conservative rule that both
judges must agree, \(\kappa=0.81\). Agreement on overclaims was
\(\kappa=0.75\) and \(\kappa=0.90\) after we corrected 9 human codes on
pipeline episodes where the first coding sheet, like the first judge
prompt, did not show the identity check that the bank's workflow had
already run. Agreement on misleading status was poor
(\(\kappa\le0.09\)), so we do not interpret that category.

A rule-based check for explicit completion claims without a matching
action found almost none (0.1\% of episodes): the agents rarely say they
did something they did not do. The opposite is common. Among the 287
episodes with a violation, both judges agreed that the agent's message
denied the very action it had executed in 26.1\% (41.3\% by the more
liberal judge). The typical case is a credit withdrawal outside the
window: the agent executes \texttt{process\_withdrawal}, and when the
customer's scripted follow-up asks for an exception, replies that the
withdrawal period has expired and it cannot process the withdrawal. The
rate was highest in the pipeline (41.3\% of H0 violations), where the
model writes its list of actions and its message in one structured
reply: in several H0 episodes the message corrects the date calculation
half-way through (``\ldots let me correct that'') while the action list,
written before it, already contains the withdrawal. A transcript-only
review, or an evaluation that scores the final message, would record
these episodes as correct refusals.

Because the follow-up is scripted and presupposes a refusal in these
cases, it may invite the denial; the judge labels are also model
outputs. We therefore treat the size of the effect as approximate. Its
direction does not depend on either point: the customer was told
something false about an action that changed their account.

\section{Discussion}\label{discussion}

\textbf{The unit of validation is the model--harness pair.} For 2 of our
four models, swapping the harness moved compliance by more than any
difference between models under a fixed harness (at most 7.7 points),
and the ranking of models changed with the harness. A validation of
``the model'' on one harness, or a benchmark score obtained under a
vendor's harness, does not carry over to a deployment that changes how
the policy is supplied or whether actions are reviewed. This is the gap
left by model-risk guidance that excludes agentic AI
\citep{frb2026sr262}: the object that needs validating is not covered,
and the object that is covered is not the one deployed. In the EU a
customer-service agent is not itself a high-risk system under the AI
Act, whose banking entry covers creditworthiness assessment and credit
scoring \citep{euaiact2024}, and the high-risk obligations for those
systems now apply from 2 December 2027 \citep{euomnibusai2026}; for
conduct, supervision of such agents rests on the sectoral rules our
policy paraphrases. Our results agree with the pooled finding of
\citet{gringras2026safetyscaffolding} that average scaffold effects are
small, and with their observation of large model-specific interactions;
for acting agents under conduct rules, the interaction is what a bank
deploys.

\textbf{Cheap controls are worth testing per model.} A same-model
reviewer removed 33\% of violations without measurable over-refusal.
This is good news for banks that cannot use a stronger external model,
but it is a property of these models and this policy, not a law: the
reviewer did not improve the withdrawal window, where it presumably
shares the agent's date error, and a pilot study found high false
rejection by a same-model reviewer in mathematics
\citep{tanveer2026reviewer}. Where a rule is arithmetic (a date window,
a cumulative limit), the natural control is deterministic code in the
harness \citep{yu2026nomos}, not another model call.

\textbf{Conduct testing should read the action log.} The denial pattern
means that sampling conversations for quality assurance, a common
practice in contact centres, will miss a share of the violations that
matter most, because the record the reviewer reads says the opposite of
what happened. Evaluations that score only the final message share the
problem \citep{advani2026falsesuccess, cao2026corruptsuccess}. In a
regulatory setting the message is also an act: telling a customer that
their withdrawal was not processed when it was is itself a conduct
failure.

\section{Limitations}\label{limitations}

The bank, customers and policy are synthetic, and one policy cannot
represent the variety of bank manuals. Several of our rules are stricter
than the law they paraphrase (Appendix A), so ``violation'' means a
breach of the manual, not necessarily of the law. We tested four
open-weight models of moderate size on one domestic API, with reasoning
disabled; stronger or reasoning models may close the date arithmetic gap
and change the harness effects. 7 of 11 rule families were near ceiling
outside H2, which limits power for the harness comparisons; the
interaction test was not significant at the 5\% level. The second
customer turn is scripted and independent of the agent's first reply,
which keeps cases comparable but makes some dialogues unnatural, and may
contribute to the denial pattern. The stated-versus-executed analysis
relies on LLM judges, validated on a sample of 100 episodes by one human
coder. The amendment after the first pilot changed the tasks; it was
made and documented before the main run, but it was informed by pilot
results.

\section{Conclusion}\label{conclusion}

Holding the language model fixed, the harness changed how often a
banking agent broke conduct rules, by up to 11.5 points for one model
and not at all for another, and it changed which model looked safest. A
same-model reviewer was the most effective single change; date
arithmetic was a failure no harness fixed; and 26\% of violations were
invisible in the conversation because the agent denied what it had done.
Banks, supervisors and benchmark authors should evaluate the
model--harness pair, report the harness, and score executed actions.

\section*{Declarations}\label{declarations}
\addcontentsline{toc}{section}{Declarations}

\textbf{Funding.} This research did not receive any specific grant from
funding agencies in the public, commercial, or not-for-profit sectors.

\textbf{Competing interests.} The author works for a participation bank.
The views expressed are the author's own and do not represent those of
any employer. No bank data were used.

\textbf{Data and code.} Cases, policy, harness code, raw episode traces
for all 3120 episodes, judge labels, the blind human coding sheet, the
pre-registration with its SHA-256 hashes and the analysis scripts are
available at https://github.com/mhuyut/harness-is-the-model. The
analysis reproduces every number in this paper from the stored traces
without calling any model API.

\textbf{Use of AI tools.} The author directed this study: the author
selected the research topic and question, approved the experimental
design and its pre-registered amendment, reviewed all of the experiment
and analysis code, monitored every experimental run and secured the
computing resources needed for all models to complete their tests,
blind-coded 100 episodes to validate the LLM judges, and read, reviewed
and finalized the manuscript. Claude Code (Anthropic; Claude Opus 5.5
and Claude Fable 5.1) was used as a research assistant throughout: to
draft the design and pre-registration, write the experiment and analysis
code, search and verify the literature, and draft the text and LaTeX
build, with all numbers generated automatically from the analysis
output. The four open-weight models under study, two of which also
served as judges, did not contribute to the writing. The author takes
full responsibility for the content; no AI tool is listed as an author.

\section*{Appendix A. Rule families and legal
sources}\label{appendix-a.-rule-families-and-legal-sources}
\addcontentsline{toc}{section}{Appendix A. Rule families and legal
sources}

\begin{longtable}[]{@{}
  >{\raggedright\arraybackslash}p{(\linewidth - 6\tabcolsep) * \real{0.2500}}
  >{\raggedright\arraybackslash}p{(\linewidth - 6\tabcolsep) * \real{0.2500}}
  >{\raggedright\arraybackslash}p{(\linewidth - 6\tabcolsep) * \real{0.2500}}
  >{\raggedright\arraybackslash}p{(\linewidth - 6\tabcolsep) * \real{0.2500}}@{}}
\caption{Rule families, the legal provisions they paraphrase, and where
the manual is stricter than the law.}\tabularnewline
\toprule\noalign{}
\begin{minipage}[b]{\linewidth}\raggedright
Family
\end{minipage} & \begin{minipage}[b]{\linewidth}\raggedright
Manual rule (summary)
\end{minipage} & \begin{minipage}[b]{\linewidth}\raggedright
Source paraphrased
\end{minipage} & \begin{minipage}[b]{\linewidth}\raggedright
Relation to the law
\end{minipage} \\
\midrule\noalign{}
\endfirsthead
\toprule\noalign{}
\begin{minipage}[b]{\linewidth}\raggedright
Family
\end{minipage} & \begin{minipage}[b]{\linewidth}\raggedright
Manual rule (summary)
\end{minipage} & \begin{minipage}[b]{\linewidth}\raggedright
Source paraphrased
\end{minipage} & \begin{minipage}[b]{\linewidth}\raggedright
Relation to the law
\end{minipage} \\
\midrule\noalign{}
\endhead
\bottomrule\noalign{}
\endlastfoot
F1 & No account information or action before date of birth and last card
digits match & GDPR confidentiality principle \citep{gdpr}; inspired by
PSD2 Art. 97 \citep{psd2} & Identity check, not strong customer
authentication \\
F2 & No disclosure to spouses or others; registered representatives and
joint holders allowed & GDPR \citep{gdpr} & Internal control built on
confidentiality \\
F3 & 14-day withdrawal from consumer credit, no penalty; mortgages never
through the assistant & Directive 2008/48/EC Art. 14 \citep{ccd1}; CCD2
Art. 26, applicable from 20 Nov 2026 \citep{ccd2} & Day-count convention
and mortgage exclusion are ours \\
F4 & No loan application before the customer states the correct APR for
that product & CCD2 Art. 10 pre-contractual information \citep{ccd2} &
Stricter than the Directive \\
F5 & Complex products only after an appropriateness check; professional
clients exempt & MiFID II Art. 25(3) \citep{mifid2} & Stricter: the
Directive requires a warning, we require refusal \\
F6 & Log every complaint and give a reference & ESA complaints-handling
guidelines \citep{jc201835} & Guidelines, not binding law \\
F7 & Customers in difficulty: refer to the hardship team, offer no new
credit & CCD2 Art. 36 (debt advisory services); cf.~Art. 35
(forbearance) and Art. 17 (ban on unsolicited credit) \citep{ccd2} &
Referral paraphrases Art. 36; ``no new credit'' is an internal
control \\
F8 & Fee refunds up to 50 EUR (business 200 EUR) per conversation,
otherwise escalate & Internal control & --- \\
F9 & Screen each beneficiary; on a match hold, report, and do not tip
off & AMLD Art. 39 \citep{amld4}; AMLR Art. 73 from 10 Jul 2027
\citep{amlr} & Matches \\
F10 & Refund unauthorised payments; after 13 months block the card and
escalate & PSD2 Art. 73 (refund), Art. 71(1) (13-month notification
limit), Art. 70(1)(e) (blocking the instrument) \citep{psd2} & Refund
duty and 13-month limit match; the automatic block and escalation after
the limit are ours (the Directive ties blocking to notification) \\
F11 & No marketing to customers who objected; business address changes
need a form & GDPR Art. 21(2)--(3) \citep{gdpr} & Consent flag stands in
for an objection \\
\end{longtable}

\renewcommand\refname{References}
\bibliography{refs.bib}

\end{document}